\documentclass{article}
\PassOptionsToPackage{numbers,compress}{natbib}

\usepackage[final,main]{neurips_2025}
\usepackage[utf8]{inputenc}
\usepackage[T1]{fontenc}
\usepackage{hyperref}
\usepackage{url}
\usepackage{booktabs}
\usepackage{amsfonts}
\usepackage{amsmath}
\usepackage{amsthm}
\usepackage{amssymb}
\usepackage{nicefrac}
\usepackage{microtype}
\usepackage{lipsum}
\usepackage{xcolor}
\usepackage{graphicx}
\usepackage{multicol}
\usepackage{tikz}
\usepackage{float}
\usepackage{placeins}
\usepackage{caption}
\usepackage{threeparttable}

\newcommand{\token}[1]{\texttt{<|#1|>}}

\graphicspath{ {./graphics/} }

\title{TTS-1 Technical Report}

\begin{document}
\maketitle

\vspace{-2cm}
\begin{center}
  \includegraphics[width=0.2\textwidth]{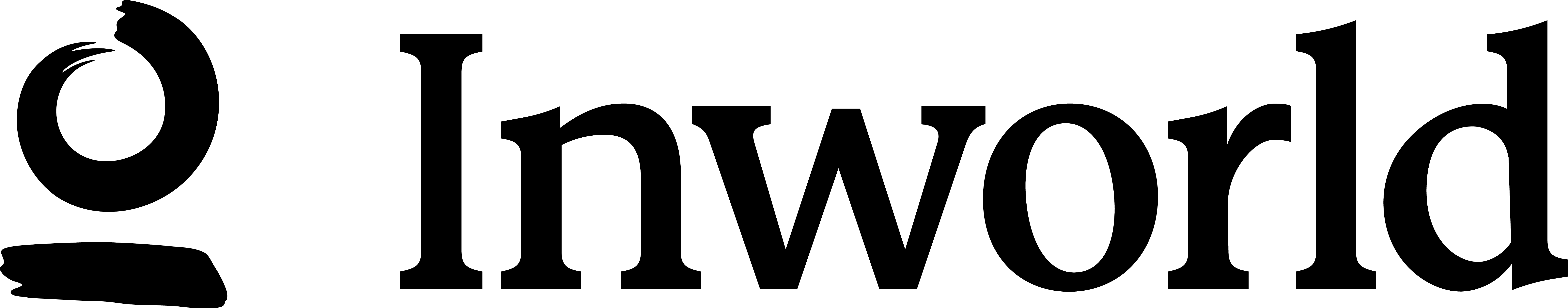}
\end{center}
\vspace{0.1cm}

\begin{abstract}
  We introduce Inworld TTS-1, a set of two Transformer-based autoregressive text-to-speech (TTS) models.
  Our largest model, \textbf{TTS-1-Max}, has 8.8B parameters and is designed for utmost quality and expressiveness in demanding applications.
  \textbf{TTS-1} is our most efficient model, with 1.6B parameters, built for real-time speech synthesis and on-device use cases.
  By scaling train-time compute and applying a sequential process of pre-training, fine-tuning, and RL-alignment of the speech-language model
  (SpeechLM) component, both models achieve state-of-the-art performance on a variety of benchmarks, demonstrating
  exceptional quality relying purely on in-context learning of the speaker's voice.
  Inworld TTS-1 and TTS-1-Max can generate high-resolution 48~kHz speech with low latency, and support 11 languages with
  fine-grained emotional control and non-verbal vocalizations through audio markups. We additionally open-source our
  training and modeling code under an MIT license.

  \vspace{0.1in}

  \textbf{Code:} \url{https://github.com/inworld-ai/tts} \\
  \textbf{Examples:} \url{https://inworld-ai.github.io/tts} \\
  \textbf{Playground:} \url{https://inworld.ai/tts}

\end{abstract}

\section{Introduction}

Recent advancements in deep learning and the proliferation of large-scale audio datasets \citep{galvez2021people, li2023yodas, he2024emilia} have propelled text-to-speech (TTS) synthesis from multi-stage pipelines \citep{li2023styletts, ren2019fastspeech} to end-to-end generative systems \citep{betker2023better, kim2021conditional, le2023voicebox}.
The latest paradigm leverages large language models (LLMs) \citep{radford2019language, hoffmann2022training} as powerful speech-language models (SpeechLMs), using neural audio codecs as tokenizers to generate highly naturalistic speech from text \citep{wang2023neural, zhang2025minimax, anastassiou2024seed, du2024cosyvoice, du2024cosyvoice2}.
Despite this progress, many existing models struggle to meet the demands of real-world applications, often lacking high-fidelity output (e.g., 48~kHz), robust multilingual support, reliable real-time streaming capabilities, or suffering from synthesis artifacts.

This paper introduces Inworld TTS-1 and TTS-1-Max, two generative speech models designed to bridge this gap.
Our models, based on 1B and 8B parameter LLaMA backbones respectively, achieve state-of-the-art speech quality and control through a systematic training methodology and architectural innovations.
We demonstrate that a sequential process of pre-training, supervised fine-tuning (SFT), and reinforcement learning (RL) alignment is critical for developing high-performance TTS systems. Our key contributions are as follows:

\begin{itemize}
  \item \textbf{A three-stage training framework for SpeechLMs.} We propose a robust pipeline consisting of (1) large-scale pre-training on over 1M hours of raw audio mixed with text data \citep{weber2024redpajama, laion-oig} to build a strong foundational model; (2) supervised fine-tuning on 200k hours of high-quality, filtered audio-text pairs; and (3) reinforcement learning alignment using Group Relative Policy Optimization (GRPO) \citep{shao2402deepseekmath} to fine-tune the model against perceptual quality metrics and reduce hallucinations.
  \item \textbf{A high-resolution audio codec for 48~kHz speech synthesis.} We develop a novel audio codec built on top of the X-codec2 \citep{ye2025llasa} architecture with a super-resolution module to natively generate 48~kHz audio. We introduce an root mean-square (RMS) loudness loss term during training to ensure volume consistency, a critical factor for streaming applications.
  \item \textbf{An extensible reinforcement learning framework for speech quality.} We adapt GRPO for TTS alignment. We design a composite reward function combining word error rate (WER), speaker similarity (SIM) \citep{chen2022wavlm}, and DNSMOS scores \citep{reddy2022dnsmos}. The framework is modular, allowing for the integration of further reward signals like prosody or emotion consistency.
  \item \textbf{Expressive and controllable speech synthesis.} We enable fine-grained control over non-verbal vocalizations and speaking styles through textual \textit{audio markups}. We show that pairing neutral and stylized utterances from the same speaker during a LoRA-based \citep{hu2022lora} fine-tuning phase is an effective strategy for teaching the model stylistic control while preserving speaker identity.
  \item \textbf{Efficient and robust streaming inference.} We detail a low-latency streaming pipeline that employs novel techniques, including context-aware decoding and concatenation at non-voicing regions, to ensure seamless and high-quality audio delivery in real-time scenarios.
\end{itemize}

Our models generate high-fidelity 48~kHz speech, support 11 languages, and offer fine-grained emotional control through in-context learning from short reference audio clips. Through extensive evaluations, we demonstrate their superior performance and practical utility for a wide range of applications, from interactive assistants to content creation. To facilitate further research and development in the community, we open-source our training, modeling, and benchmarking code under a permissive license.

\section{Architecture}

\begin{figure}[htbp]
  \centering
  \includegraphics[width=0.95\linewidth]{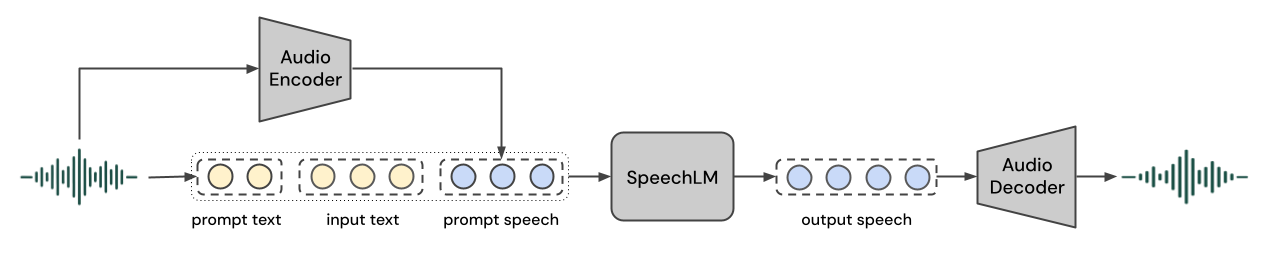}
  \caption[]{The architecture of Inworld TTS-1. The audio encoder tokenizes a reference audio into a sequence of discrete audio tokens. These tokens are concatenated with the tokenized reference text and the text to be synthesized to form a prompt for the SpeechLM. The SpeechLM autoregressively generates audio tokens, which are then converted back into a 48~kHz waveform by the audio decoder.}
  \label{fig:architecture}
\end{figure}

Inworld TTS-1 and TTS-1-Max are built on the same architecture as shown in Figure \ref{fig:architecture}: they use the same audio encoder and decoder components and only differ in the size of the SpeechLM backbone used. With all inference components combined, the models have 1.6B and 8.8B parameters, respectively. Below we outline key aspects of both components.

\subsection{Audio Codec}

Audio codec selection plays a critical role in building a high-quality TTS model. We opted for the
codec architecture of X-codec2 \citep{ye2025llasa}, which merges both acoustic and semantic information of
the encoded audio into a single codebook of 65536 tokens and generates 50 tokens per second of audio.

The selection of this codec architecture was motivated by several key factors:

\begin{itemize}
  \item \textbf{Streaming inference}. The architectural simplicity and 1D causal structure of the audio tokens enable efficient streaming inference with minimal communication overhead.
  \item \textbf{Open-source implementation}. The availability of source code facilitated training the codec from scratch on large-scale datasets and enabled adaptation for high-resolution audio processing (the original implementation was limited to 16~kHz audio).
  \item \textbf{Computational efficiency}. The compact one-dimensional codebook structure enables efficient storage and processing of tokenized audio data. For instance, encoding 1 hour of raw 48~kHz mono audio requires $\sim$365MB of storage, while the tokenized representation from a codebook of 65536 tokens requires only $\sim$0.19MB using \texttt{uint16} data types.

\end{itemize}

To support high-resolution speech synthesis, we augmented the original X-codec2 decoder with a super-resolution module.
The baseline X-codec2 decoder architecture employs a backbone model comprising ResNet blocks \citep{he2016deep} and transformer layers to predict acoustic features from audio tokens.
These predicted acoustic features are subsequently converted to an audio waveform via an inverse short-time Fourier transform (iSTFT).

The temporal resolution of the generated audio is determined by two factors: the resolution of the predicted acoustic features and the hop length parameter of the iSTFT operator. In the original implementation, for one second of audio, the decoder's backbone model processes 50 input tokens and applies iSTFT with a hop length of 320 samples, yielding 16~kHz audio output.
Our super-resolution module extends this architecture through interleaved strided 1D transposed convolutional layers and ResNet blocks.
The transposed convolution layers increase the temporal resolution of predicted acoustic features by predetermined stride factors.
When combined with appropriately adjusted iSTFT hop lengths, this module enables audio generation with higher sample rates while introducing minimal computational overhead.

Table \ref{tab:codec_params} provides comprehensive details regarding the stride configurations and corresponding iSTFT hop lengths employed in our implementation.

\begin{table}[htbp]
  \centering
  \begin{tabular}{lcccc}
    \toprule
    \textbf{Component} & \textbf{Sample Rate} & \textbf{Strides} & \textbf{Hop length} & \textbf{Parameter Count, M} \\
    \midrule
    Audio Encoder      & 16~kHz     &  N/A & 320 & 55.28 \\
    Audio Decoder      & 16~kHz     &  1 & 320 & 187.00 \\
    Audio Decoder      & 24~kHz     &  1 & 480 & 187.65 \\
    Audio Decoder      & 48~kHz     & 3, 2 & 160 & 193.43 \\
    \bottomrule
  \end{tabular}
  \vspace{0.1cm}
  \caption{Trainable parameter counts for the audio encoder and decoder.}
  \label{tab:codec_params}
\end{table}

Notably, our proposed decoder uses a significantly larger hop length compared to other iSTFT-based audio generation models. For example, \citet{du2024cosyvoice2} use a hop length of 4 for 24~kHz audio generation, while we empirically found that for 48~kHz audio generation our model yields better speech quality.

\subsection{SpeechLM}

TTS-1 and TTS-1-Max employ LLaMA-3.2-1B and LLaMA-3.1-8B \citep{grattafiori2024llama} as their respective SpeechLM backbones.
The vocabulary of each model was expanded from 128256 to 193856 tokens, incorporating 65536 audio tokens and 29 special tokens.
The resulting vocabulary size was padded to be a multiple of 32 for computational efficiency \citep{nvidia_tc_matmul}.
Notably, a vocabulary size of 193800 was used during an intermediate pre-training phase before the final requirements for instruction-following were established.

The embeddings for new vocabulary tokens were initialized by sampling from a multivariate normal distribution with the mean and covariance of the original embedding matrix \citep{hewitt2021initializing}.

\section{Training Methodology}

\subsection{Data}
\label{sec:data}

The training corpus for our models comprises a large-scale, multilingual audio-text dataset curated from a combination of publicly available sources and licensed third-party providers.
This corpus is partitioned into two main subsets for pre-training and supervised fine-tuning, respectively.

\begin{figure}[H]
  \centering
  \includegraphics[width=0.8\linewidth]{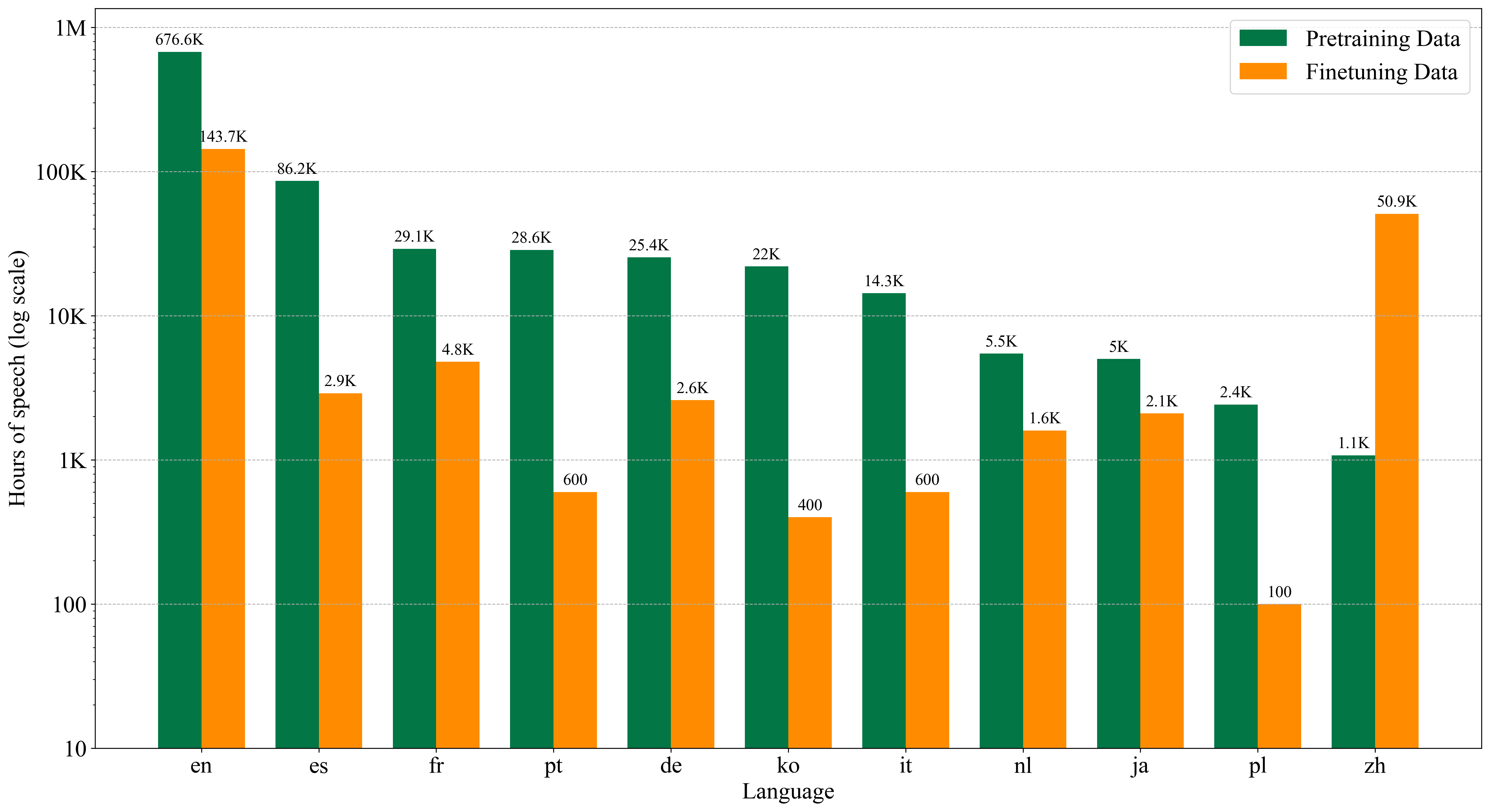}
  \caption[]{Speech data quantities per language and dataset.}
  \label{fig:data_language_distribution}
\end{figure}

\textbf{Pre-training.} The pre-training dataset contains $\sim$1 million hours of raw audio.
To enhance model robustness against diverse, real-world acoustic environments, this set is augmented with $\sim$30000 hours of non-speech audio, including environmental sounds, background noise, and multi-speaker babble.
The linguistic composition of the speech portion is detailed in Figure \ref{fig:data_language_distribution}. Additionally, an initial $\sim$15000-hour high-quality subset from our fine-tuning data was incorporated into the pre-training mixture to bootstrap the model's text-to-speech alignment capabilities.

\textbf{Fine-tuning.} The supervised fine-tuning (SFT) dataset consists of $\sim$200000 hours of high-quality, transcribed audio-text pairs.
This data was carefully filtered and prepared for subsequent training stages. Further details on data processing and annotation for specific training phases are provided in the corresponding sections.

\FloatBarrier

\subsection{Audio Codec}

We follow the same methods as X-codec2 \citep{ye2025llasa} to train the audio codec.
The only notable difference is how we trained the audio decoder component, which we describe in more detail below.

\textbf{Experiments.} We train our audio codec on about 110k hours of multilingual audio data sourced from both pre-training and fine-tuning datasets described in Section \ref{sec:data}. In early experiments, we observed that the decoder's output often failed to preserve the perceived volume of the original audio, especially when decoding short or high-pitched segments. As a consequence, when these segments are concatenated during streaming, the user may perceive sudden increases or drops in volume.

To address this, we introduced an additional root mean-square (RMS) loudness loss term that penalizes discrepancies in volume between the original and decoded waveforms. RMS is a standard statistical measure used to approximate perceived loudness, making it a natural choice for enforcing consistency. The training loss is defined similarly to \citet{xin2024bigcodec}:

\begin{equation}
  \mathcal{L}_{decoder} =  \lambda_{\mathrm{disc}} \cdot \mathcal{L}_\mathrm{disc} + \mathcal{L}_{gen} + \lambda_{\mathrm{RMS}} \cdot \mathcal{L}_{RMS}
\end{equation}

\begin{equation}
  \mathcal{L}_{gen}  = \lambda_{\mathrm{adv}} \cdot \mathcal{L}_{adv} + \lambda_{\mathrm{fm}} \cdot \mathcal{L}_{fm} + \lambda_{\mathrm{mel}} \cdot \mathcal{L}_{mel}
\end{equation}

where $\mathcal{L}_{disc}$ is the discriminator loss derived from the multi-period discriminator (MPD) \citep{kong2020hifi} and the multi-scale STFT discriminator (MSD) \citep{ye2025llasa, defossez2022highfi}; $\mathcal{L}_{gen}$ is the generator loss, which includes the mel spectrogram loss, feature matching loss, and the adversarial loss \citep{kong2020hifi}. The new $\mathcal{L}_{RMS}$ term is defined as:

\begin{equation}
  \mathcal{L}_{RMS} = \mathbb{E}\left[X_{t} - \hat{X}_{t}\right]^2
\end{equation}

\begin{equation}
  X_{t} = 20 \log_{10}\left(\sqrt{\frac{1}{T} \sum_{t=1}^{T} x_t^2} + \epsilon\right)
\end{equation}

where $X_{t}$ and $\hat{X}_{t}$ are the original and generated audio samples volumes in decibels, $T$ is the
corresponding waveform array length, and $\epsilon = 10^{-5}$ is a small constant for numerical stability.
In our experiments we found that $\lambda_{\mathrm{RMS}} = 1.0$ works well for both 24~kHz and 48~kHz audio.

To achieve 48~kHz audio synthesis, up-training was performed in two stages: first with audio samples having sample rates $\geq 32$~kHz, then with additional fine-tuning on audio with native sample rates $\geq 44.1$~kHz. This progressive training strategy ensures robust high-fidelity audio generation across varied scenarios.

\textbf{Evaluation results.}
Subjective evaluation reveals that the 48~kHz decoder produces clearer audio with richer high-frequency details. This finding is consistent with the DNSMOS scores presented in Table \ref{tab:decoder_dnsmos}, where the 48~kHz decoder achieves the highest audio quality.

\begin{table}[H]
  \centering
  \begin{tabular}{lc}
    \toprule
    \textbf{Sample Rate} & \textbf{DNSMOS} \\
    \midrule
    16~kHz & 4.110 \\
    24~kHz & 4.178 \\
    48~kHz & 4.195 \\
    \bottomrule
  \end{tabular}
  \vspace{0.1cm}
  \caption{DNSMOS scores for decoder models trained with different sample rates. The evaluation is performed on the English subset of the Seed-TTS evaluation \citep{anastassiou2024seed}.}
  \label{tab:decoder_dnsmos}
\end{table}

\subsection{SpeechLM Pre-Training}

\label{sec:speechlm_pre_training}

To enable the model to precisely recreate an arbitrary speaker's voice, encompass a wide range of pitch variations, and reproduce non-verbal vocalizations, the model was pre-trained on a diverse corpus of authentic speech.

\textbf{Experiment setup.} The raw audio data underwent minimal processing and was segmented into samples with a maximum duration of 40 seconds, a constraint imposed by the audio encoder's architecture.
Each sample was demarcated by \texttt{$\token{speech\_start}$} and \texttt{$\token{speech\_end}$} special tokens to facilitate the model's ability to learn continuous speech patterns.

The audio data was supplemented with approximately 10\% text data from the RedPajama-v2 pre-training dataset \citep{weber2024redpajama}, which was filtered using simple heuristics to retain high-quality, long-form documents.
Also, following the findings of \citep{prabhumoye2023adding}, we added instruction data from LAION OIG \citep{laion-oig} as raw text tokens to the text dataset to help the model's ability to understand text.
Overall, around 20B text tokens were used for both models' pre-training. This approach was observed to prevent the degradation of text comprehension abilities during subsequent supervised fine-tuning stages, without adversely affecting speech synthesis quality.

\begin{figure}[H]
  \centering
  \includegraphics[width=0.8\linewidth]{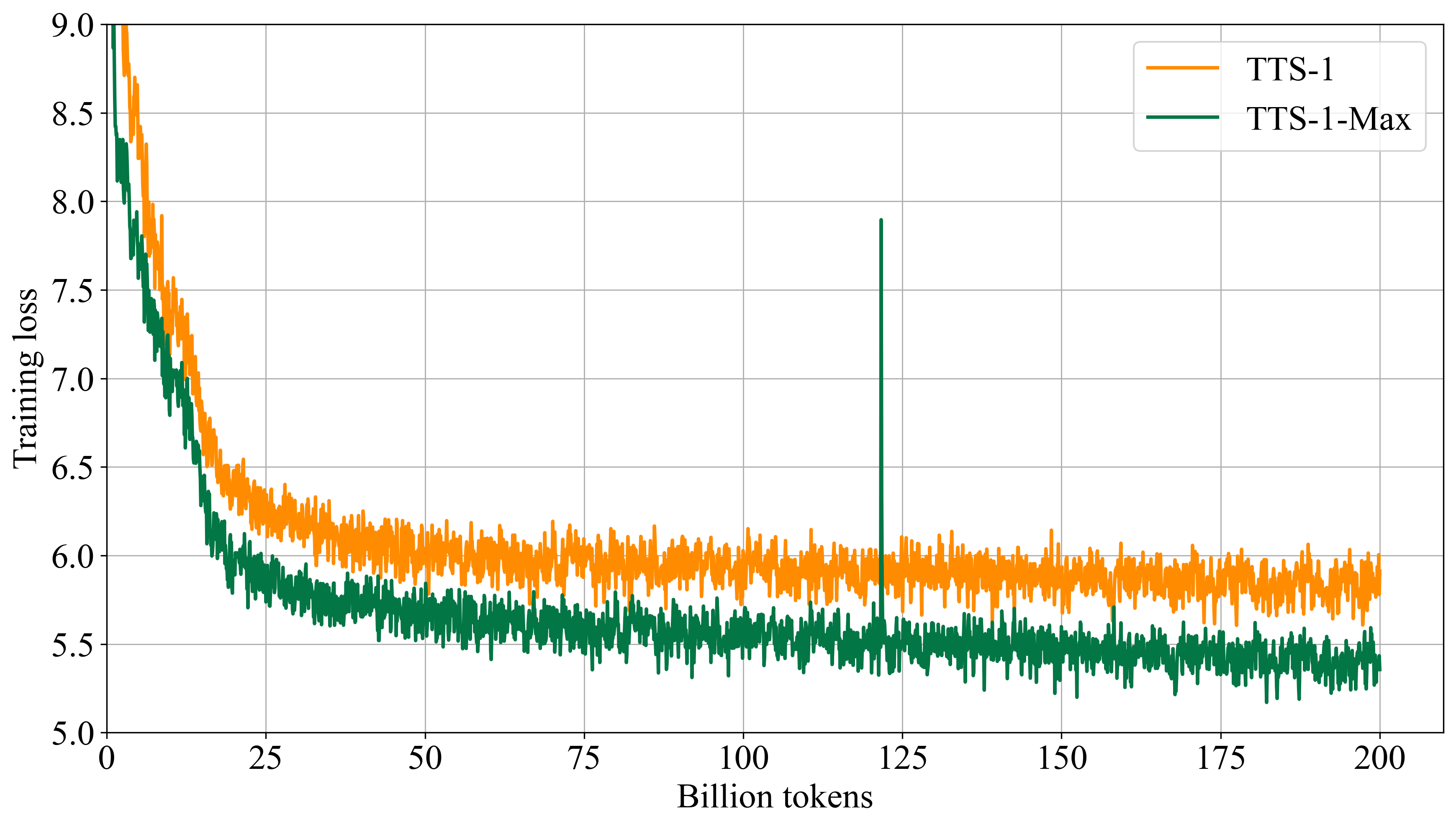}
  \caption[]{Pre-training loss for the TTS-1 and TTS-1-Max SpeechLM models.}
  \label{fig:pretraining_loss}
\end{figure}

For training we use a standard technique to perform unsupervised language modeling via next token prediction \citep{bengio2003neural} using the hyperparameters shown in Table \ref{tab:pretrain_hparams}.

\begin{table}[htbp]
  \centering
  \begin{tabular}{lcc}
    \toprule
    \textbf{Hyperparameter} & \textbf{TTS-1} & \textbf{TTS-1-Max} \\
    \midrule
    Training strategy & DDP & FSDP \\
    Per-GPU batch size & 8 & 2 \\
    Gradient accumulation steps & 1 & 4 \\
    Total number of GPUs & \multicolumn{2}{c}{32} \\
    Maximum sequence length & \multicolumn{2}{c}{2048} \\
    Optimizer & \multicolumn{2}{c}{AdamW, $\beta_1 = 0.9$, $\beta_2 = 0.95$} \\
    Learning rate (peak) & \multicolumn{2}{c}{$1.5 \times 10^{-4}$} \\
    Learning rate schedule & \multicolumn{2}{c}{Cosine decay (10\% warmup)} \\
    Weight decay & \multicolumn{2}{c}{0.1} \\
    \bottomrule
  \end{tabular}
  \vspace{0.1cm}
  \caption{Pre-training hyperparameters for TTS-1 and TTS-1-Max.}
  \label{tab:pretrain_hparams}
\end{table}

To optimize training throughput, we employed a combination of \texttt{bf16} mixed-precision training, FlashAttention-2 \citep{dao2023flashattention}, and the fused AdamW optimizer \citep{kingma2014adam}.
For distributed training, the Distributed Data Parallel (DDP) strategy \citep{li2020pytorch} was applied to the smaller TTS-1 model.
For the larger TTS-1-Max model, we adopted the Fully Sharded Data Parallel (FSDP) strategy \citep{zhao2023pytorch} to maximize GPU utilization.
Additionally, given the fixed sequence length of any given batch during pre-training we used \texttt{torch.compile()} to further optimize the training throughput.

On average we were able to achieve $\sim$46000 and $\sim$8000 tokens/sec/GPU for TTS-1 and TTS-1-Max respectively using 32 H100 GPU with 80GB of memory.
Single pre-training run for TTS-1 and TTS-1-Max took $\sim$2 and $\sim$10 days respectively.

\textbf{Evaluation results.} The performance of the pre-trained TTS-1 and TTS-1-Max models is benchmarked on a multi-lingual test set of approximately 5,000 utterances with an average duration of 10 seconds.
Voice similarity is quantified as the cosine similarity (SIM) between speaker embeddings from the generated and reference audio, extracted using a WavLM-large model fine-tuned for speaker verification \citep{chen2022wavlm}.
For our evaluation, each utterance is bisected: the first half serves as the reference audio prompt for the model to generate the subsequent half, which is then compared against the ground truth.
Speech is synthesized by decoding the raw token distribution with $top_k=50$, $top_p=1.0$ and $T=1.0$.

\begin{figure}[htbp]
  \centering
  \includegraphics[width=0.8\linewidth]{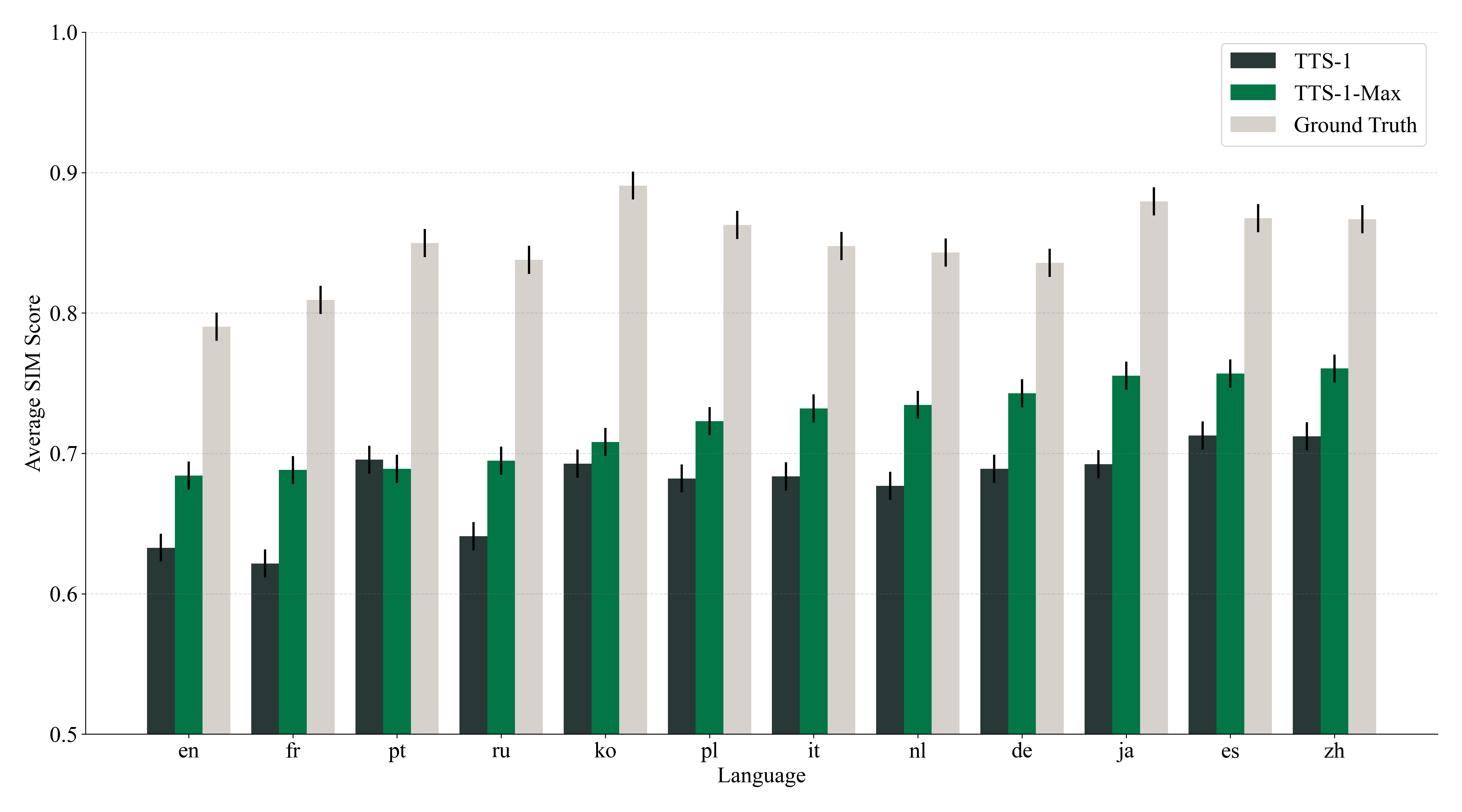}
  \caption[]{Performance of TTS-1 and TTS-1-Max pre-trained SpeechLM checkpoints on the test set.
  Cross-language results are not directly comparable as each language subset contains distinct utterances.}
  \label{fig:pretraining_sim_eval}
\end{figure}

The results, presented in Figure \ref{fig:pretraining_sim_eval}, indicate that the overall performance correlates with the training loss difference between both models that is depicted in Figure \ref{fig:pretraining_loss}.
Notably, the improvements in prompt continuation similarity are more pronounced for the larger TTS-1-Max model compared to TTS-1.

\subsection{SpeechLM Supervised Fine-Tuning}

For supervised fine-tuning (SFT), we adopt a standard autoregressive training objective, consistent with prior work
in speech synthesis \citep{anastassiou2024seed, du2024cosyvoice, du2024cosyvoice2, ye2025llasa}.
The model is trained to predict each audio token conditioned on the input text prompt and the preceding ground-truth audio tokens.
This is achieved by minimizing the negative log-likelihood of the target sequence, defined as:

\begin{equation}
  \mathcal{L}_{\text{SFT}} = - \sum_{s=1}^{S} \log P(y_s | x_1, \ldots, x_T, y_1, \ldots, y_{s-1})
\end{equation}

where $\{x_1, \ldots, x_T\}$ represents the sequence of input text tokens and $\{y_1, \ldots, y_S\}$ is the sequence of target audio tokens.
During the training the input sequence is constructed as follows:

\begin{equation}
  \left[\token{begin\_of\_text}, x_{1}, \ldots, x_{T}, \token{speech\_start}, y_{1}, \ldots, y_{S}, \token{speech\_end}\right]
\end{equation}

\textbf{Experiment setup.}
To enhance the quality of the generated speech, the audio data was systematically pre-processed.
We annotated the dataset with DNSMOS scores \citep{reddy2022dnsmos} to quantify audio quality and measured characters per second (CPS) to determine talking speed.
A multi-stage filtering process was then applied. Inspired by \citep{lacombe-etal-2024-parler-tts}, we first discarded the 20\% of samples with the lowest DNSMOS scores, which correspond to the low-quality tail of the distribution.
Next, we computed language-specific CPS distributions and filtered out the fastest 5\% and slowest 5\% of samples, as these outliers often contain malformed audio or transcription errors.
Finally, we applied text-based heuristics to remove samples with low-quality transcriptions, such as those containing only punctuation, or other non-speech content.
This filtering pipeline yielded a final dataset of approximately 200000 hours of high-quality audio.

As an initial checkpoint for both TTS-1 and TTS-1-Max, we used the audio pre-trained SpeechLM checkpoints from the previous stage (Section \ref{sec:speechlm_pre_training}).
We found that initializing the SFT learning rate to the final learning rate from pre-training was crucial for achieving optimal speech quality. The other hyperparameters for SFT are detailed in Table \ref{tab:sft_hparams}.

\begin{table}[htbp]
  \centering
  \begin{tabular}{lcc}
    \toprule
    \textbf{Hyperparameter} & \textbf{TTS-1} & \textbf{TTS-1-Max} \\
    \midrule
    Training strategy & DDP & Deepspeed Stage 2 \\
    Per-GPU batch size & 4 & 2 \\
    Gradient accumulation steps & 1 & 2 \\
    Total number of GPUs & \multicolumn{2}{c}{32} \\
    Maximum sequence length & \multicolumn{2}{c}{2048} \\
    Optimizer & \multicolumn{2}{c}{AdamW, $\beta_1 = 0.9$, $\beta_2 = 0.95$} \\
    Learning rate (peak) & \multicolumn{2}{c}{$1.5 \times 10^{-5}$} \\
    Learning rate schedule & \multicolumn{2}{c}{Cosine decay (5\% warmup)} \\
    Weight decay & \multicolumn{2}{c}{0.1} \\
    \bottomrule
  \end{tabular}
  \vspace{0.1cm}
  \caption{SFT hyperparameters for TTS-1 and TTS-1-Max.}
  \label{tab:sft_hparams}
\end{table}

Throughout the SFT training, the audio codec was loaded on each GPU to periodically perform inference to generate
audio samples for quality evaluations. This imposed an additional memory overhead of $\sim$4GB per GPU; therefore, the
FSDP setup requiring more memory for a larger model was replaced with Deepspeed Stage 2 \citep{rajbhandari2020zero}.
Also, given the variable maximum sequence length of any given batch during SFT, \texttt{torch.compile()} was not applicable, and the final
training throughput was approximately $\sim$18000 and $\sim$4800 tokens/sec/GPU for TTS-1 and TTS-1-Max, respectively, using the same training cluster as for pre-training.

\textbf{Challenges with mixed-objective fine-tuning.} We investigated the effect of incorporating text-based instruction-following data during SFT to improve the model's interpretation of nuanced prompts.
However, this approach led to a notable degradation in synthesis quality, with the model often failing to generate speech reliably.
This occurred despite observing no adverse impact on the training loss for the audio portion of the mixed dataset.

\begin{figure}[H]
  \centering
  \includegraphics[width=0.8\linewidth]{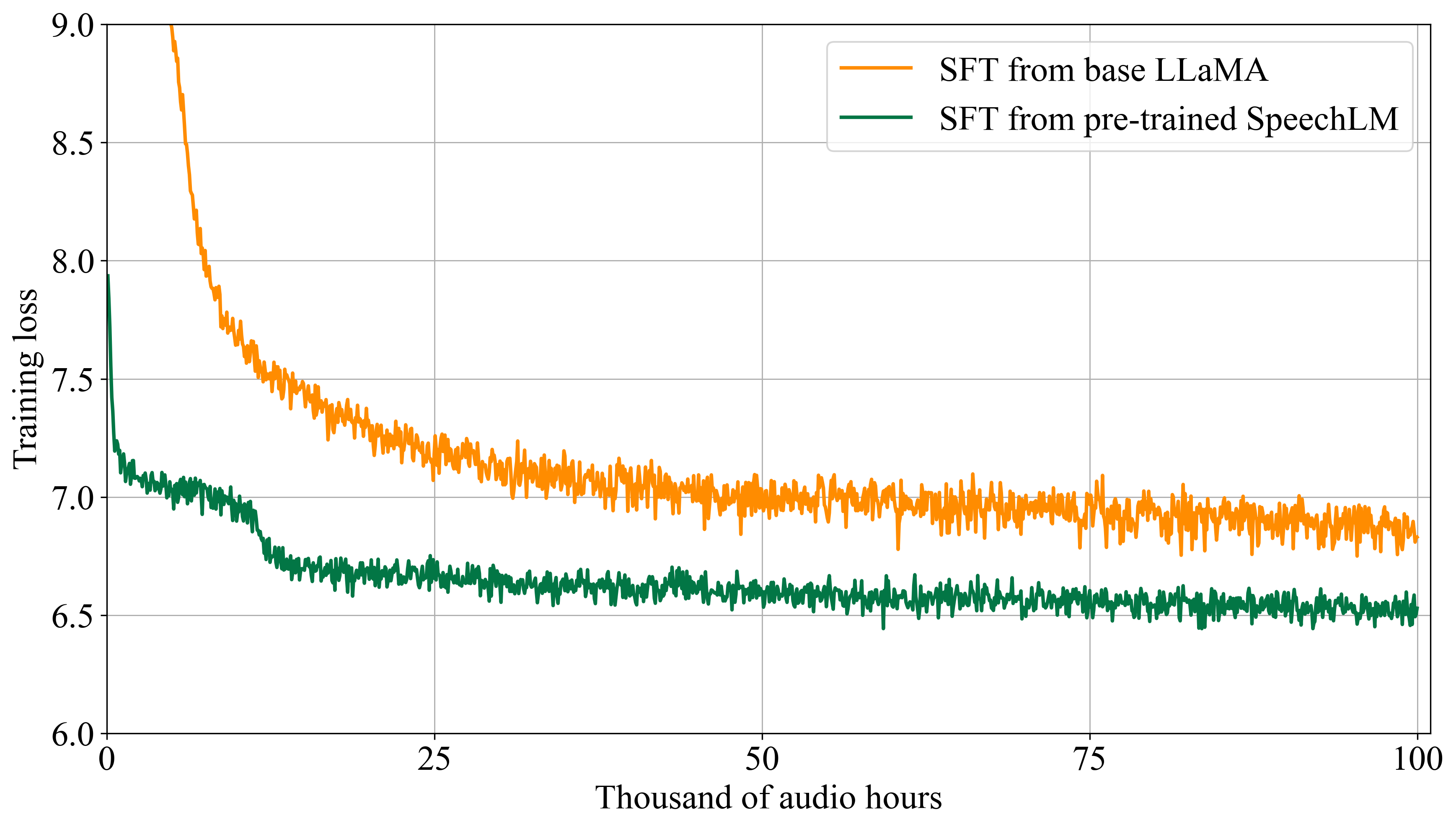}
  \caption[]{Ablation study on the impact of audio pre-training for SFT of TTS-1.
    The plot compares SFT loss curves for models initialized from the base LLaMA-3.2-1B checkpoint versus an audio pre-trained version.
  Both were trained on $\sim$100k audio hours under identical settings.}
  \label{fig:1b_sft_pretraining_impact_ablation}
\end{figure}

\textbf{Impact of pre-training.} The use of a pre-trained SpeechLM as the starting point for SFT proved to be crucial for achieving high-quality results.
As shown in Figure \ref{fig:1b_sft_pretraining_impact_ablation}, an ablation study on a 100k-hour multilingual data subset demonstrates that fine-tuning from an audio pre-trained checkpoint yields a lower training loss compared to starting from the base LLaMA-3.2-1B-Instruct checkpoint.

Additionally, our internal benchmark evaluation shows that pre-training provides substantial improvements in objective metrics: WER improved by approximately 15\% and speaker similarity improved by approximately 3\% compared to the version without pre-training.

The SFT quality evaluation results are presented in Section \ref{sec:quality_evaluation}.

\subsection{SpeechLM RL-Alignment}

While supervised fine-tuning (SFT) enables the model to generate coherent speech, it inherently optimizes for maximum likelihood estimation on the training data, which may not align with human preferences for speech quality.
Speech synthesis presents unique challenges where traditional automatic metrics used in SFT training fail to capture perceptual qualities such as naturalness, expressiveness, and speaker similarity.
Moreover, autoregressive speech generation is prone to hallucinations—generating spurious audio tokens that result in artifacts like clicks, pops, or unnatural vocalizations that degrade the listening experience.

Recent advances in text-to-speech research have demonstrated that reinforcement learning (RL) techniques can significantly improve the perceptual quality of synthesized audio \citep{anastassiou2024seed,du2025cosyvoice,chen2024enhancing}.
Furthermore, the success of models such as those in the DeepSeek family \citep{liu2024deepseek,shao2402deepseekmath,bi2024deepseek} have established RL as a dominant paradigm in large language model development.
Techniques such as Direct Preference Optimization (DPO) \citep{rafailov2023direct} and Group Relative Policy Optimization (GRPO) \citep{shao2402deepseekmath} have shown remarkable success in steering model behavior toward human-preferred outcomes, improving both the reliability and usefulness of generated content through iterative feedback mechanisms.
Motivated by these developments, we adapt these methodologies to the domain of speech generation, where the primary challenge involves crafting reward signals that effectively capture the complex, subjective aspects of audio quality.

To address these limitations, we employ reinforcement learning (RL) to align the SpeechLM with human preferences for high-quality speech synthesis.
Specifically, we use Group Relative Policy Optimization (GRPO) \citep{shao2402deepseekmath}, a variant of Proximal Policy Optimization (PPO) that is particularly well-suited for language model alignment tasks.

\textbf{GRPO formulation.} GRPO optimizes the SpeechLM by encouraging it to generate responses that maximize a given reward function while staying close to the original supervised fine-tuned model.
It achieves this by updating the model based on an advantage estimate, which measures how much better a generated response is compared to the average response in a batch.
To ensure training stability, policy updates are clipped to prevent large deviations from the reference model.
Furthermore, a KL divergence penalty is used as a regularization component that constrains the updated policy from diverging excessively from the reference policy,
preventing the model from generating degenerate outputs that could arise from unconstrained optimization.

The key innovation in GRPO is the computation of advantages using group-based comparisons rather than individual rewards.
For a batch of $G$ responses $\{o_1, \ldots, o_G\}$ generated for the same query $q$, the advantage for response $o_i$ is computed as:
\begin{equation}
  \hat{A}_{i,t} = r_i - \text{mean}(\mathbf{r})
\end{equation}

where $r_i$ is the reward for response $o_i$, and $\mathbf{r} = \{r_1, \ldots, r_G\}$ is the vector of rewards for all responses in the group.
As demonstrated in \citep{liu2025understanding}, scaling rewards can introduce a question-level difficulty bias.
To address this, we disable reward scaling in GRPO and by default use unscaled, mean-centered advantages for more robust training.

\textbf{Reward pipeline.} To compute the reward signals, each output sequence of audio tokens from the SpeechLM is processed through a multi-stage evaluation pipeline.
First, the generated audio token sequences are passed through the audio decoder to reconstruct the corresponding 48~kHz waveforms.
These decoded audio samples are then fed into specialized evaluation models to compute the individual reward components: an automatic speech recognition (ASR) model that transcribes the audio to compute WER scores, a speaker verification model extracts embeddings to calculate similarity scores with the reference audio,
and a DNSMOS model assesses the perceptual quality of the generated speech.
This evaluation pipeline ensures that the reward signals accurately reflect the quality of the final audio output rather than just the token-level predictions.

This group-relative approach reduces variance in advantage estimation and provides more stable training compared to absolute reward-based methods.

\textbf{Reward function design.} Our reward function $R(p, c)$ combines three key components to capture different aspects of speech quality, where $p$ represents the prompt input and $c$ represents the generated completion audio:
\begin{equation}
  R(p, c) = \alpha R_{wer}(c) + \beta R_{similarity}(p, c) + \gamma R_{dnsmos}(c)
\end{equation}

where each component is normalized to the $[0, 1]$ range and addresses specific aspects of speech synthesis quality:

\textbf{WER reward.} $R_{wer}$ represents the word error rate reward, computed by transcribing the generated audio using Whisper-large-v3 model \citep{radford2023robust} and comparing it against the target text. The raw WER score is normalized using an exponential decay function to map it to the $(0, 1]$ range:
\begin{equation}
  R_{wer}(c) = \exp(-k \cdot \text{WER}(c))
\end{equation}
where $k = 2.5$ is a scaling parameter that controls the sensitivity to WER values. For languages in the character error rate (CER) evaluation list, character error rate is used instead of word error rate.

\textbf{Similarity reward.} $R_{similarity}$ evaluates speaker similarity between generated and reference audio using deep speaker verification models based on WavLM-large fine-tuned for speaker verification \citep{chen2022wavlm}. The cosine similarity score is computed between speaker embeddings and is already in the range $[-1, 1]$, normalized as:
\begin{equation}
  R_{similarity}(p, c) = \frac{\text{CosSim}(\text{embed}(p), \text{embed}(c)) + 1}{2}
\end{equation}

\textbf{DNSMOS reward.} $R_{dnsmos}$ evaluates perceptual speech quality using the Deep Noise Suppression Mean Opinion Score (DNSMOS) \citep{reddy2022dnsmos}, whose results are strongly aligned with how humans judge audio quality. The DNSMOS score is already in the range $[1, 5]$ and is normalized as:
\begin{equation}
  R_{dnsmos}(c) = \frac{\text{DNSMOS}(c) - 1}{4}
\end{equation}

The coefficients $\alpha, \beta, \gamma$ are hyperparameters that balance the relative importance of each component, with careful tuning required to achieve optimal trade-offs between speech quality, speaker fidelity, and content accuracy. All reward components are designed to be maximized, with higher values indicating better performance.

\textbf{Extensibility and flexibility.} The modular design of our reward function makes it highly extensible for incorporating additional reward signals that capture various aspects of human preferences in TTS perceptual quality. The framework can easily accommodate new reward components by simply extending the linear combination:
\begin{equation}
  R(p, c) = \sum_{i} w_i R_i(p, c)
\end{equation}

where $w_i$ are the corresponding weights and $R_i$ are individual reward functions.

This flexible architecture allows for rapid experimentation with different reward combinations and enables the incorporation of domain-specific quality metrics tailored to particular applications or user preferences.

\textbf{Conditional reward activation.} An important feature of this framework is the ability to conditionally activate reward signals based on dataset annotations or prompt characteristics.
For example, speaking style rewards (e.g., $R_{style}$) are only applied to samples that contain annotated speaking style tags like \texttt{[whispering]} or \texttt{[laughing]}, while emotion consistency rewards ($R_{emotion}$) are activated only for prompts with emotional markup tags such as \texttt{[angry]} or \texttt{[sad]}.
Similarly, non-verbal token rewards ($R_{nonverbal}$) are computed only for samples containing non-verbal annotations like \texttt{[breathe]} or \texttt{[sigh]}. This conditional approach ensures that specialized reward signals are applied only to relevant training examples, preventing unnecessary computation and avoiding potential conflicts when evaluating samples that lack the corresponding annotations.

\textbf{Experiments.} We conducted two sets of experiments to evaluate the effectiveness of our GRPO-based RL alignment.
Both experiments used the SFT-trained TTS-1 model as the reference policy and were performed on a $1,000$-hour subset of our English fine-tuning data.
For each prompt, we generated eight responses and computed the rewards for each.

In the first set of experiments, we trained three separate models, each optimized for a single reward signal: WER, speaker similarity, or DNSMOS.
In the second experiment, we trained a single model using a combined reward function. For this set of experiments, we weighted WER, speaker similarity, and DNSMOS rewards equally ($\alpha=\beta=\gamma=1.0$), but it is important to note that these weights can be tuned to prioritize different optimization goals.

\begin{figure}[htbp]
  \centering
  \includegraphics[width=0.95\linewidth]{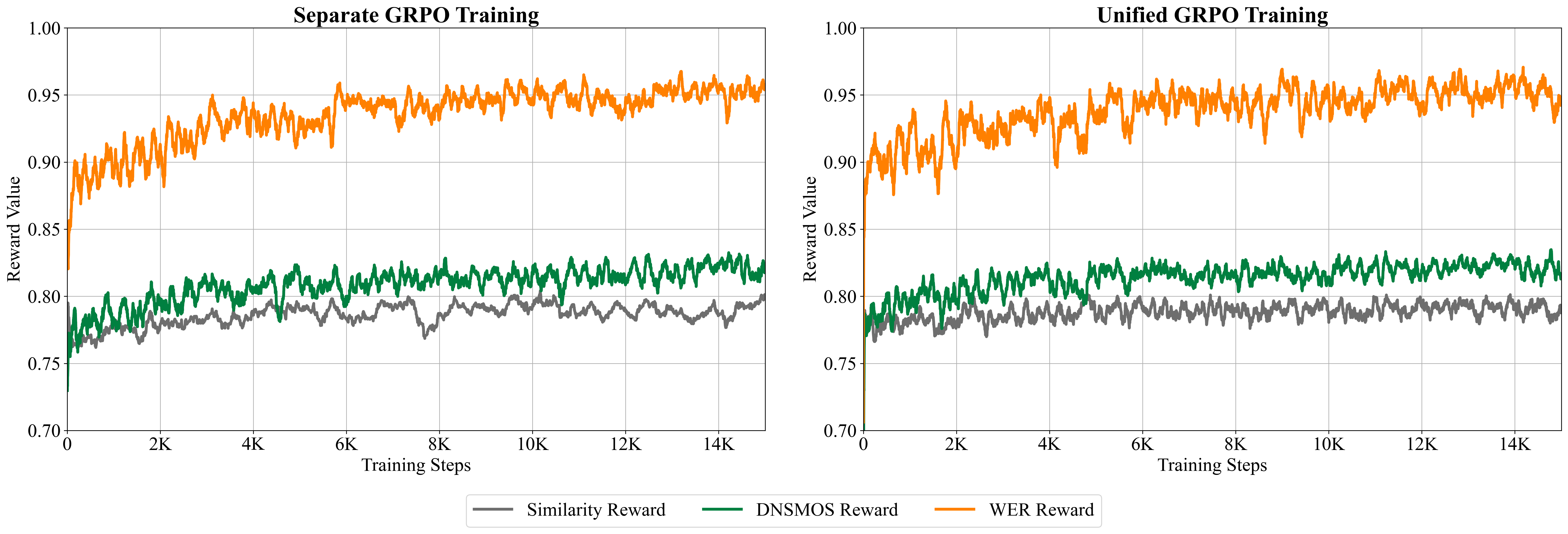}
  \caption{Training curves for GRPO experiments. Left: separate reward signal training, where each model was trained with a single reward (WER, Similarity, or DNSMOS) and evaluated on the corresponding metric. Right: unified model with combined reward function.}
  \label{fig:rlhf_training}
\end{figure}

Figure \ref{fig:rlhf_training} shows the training curves for both experimental setups. The left panel demonstrates that each individual reward signal was effective at improving its corresponding metric when trained separately. The right panel shows that the unified approach achieved a strong balance across all three metrics, outperforming the specialized models in overall quality. The combined model demonstrated significant improvements in all three dimensions compared to the SFT baseline, validating our approach of using a composite reward to align the model with multiple facets of human preference for speech quality.

\subsection{Audio Markups}

To facilitate control over speech synthesis prosody and style, independent of the reference audio, we employ embedded text annotations to guide the generation process.
These annotations, which we term \textit{audio markups}, are categorized into 8 speaking styles and 7 non-verbal vocalizations, as detailed in Table \ref{tab:audio_markup_tags}.

\begin{table}[htbp]
  \centering
  \begin{tabular}{lp{0.65\linewidth}}
    \toprule
    \textbf{Category} & \textbf{Tags} \\
    \midrule
    Speaking style & \texttt{[angry]}, \texttt{[disgusted]}, \texttt{[fearful]}, \texttt{[happy]}, \texttt{[laughing]}, \texttt{[sad]}, \texttt{[surprised]}, \texttt{[whispering]} \\
    \addlinespace
    Non-verbals & \texttt{[breathe]}, \texttt{[clear\_throat]}, \texttt{[cough]}, \texttt{[cry]}, \texttt{[laugh]}, \texttt{[sigh]}, \texttt{[yawn]} \\
    \bottomrule
  \end{tabular}
  \vspace{0.1cm}
  \caption{Types of audio markup tags used for conditioning speech synthesis.}
  \label{tab:audio_markup_tags}
\end{table}

\textbf{Dataset construction.} Non-verbal vocalizations are relatively straightforward to incorporate due to their temporally localized nature.
The SpeechLM learns to generate these sounds reliably through their direct annotation in the input prompt.

Style control tags present a greater challenge, as the model must maintain high speaker similarity to the reference audio while simultaneously exhibiting the nuanced desired style, which typically corresponds to a specific emotion.
This emotion influences the entire utterance, affecting its pace, prosody, tone, and volume.

In initial experiments, wherein individual examples with prepended style control tags were introduced during SFT, we observed that the model failed to utilize these tags, producing no discernible effect on the output.
We hypothesized that this failure stems from the audio codec's design, which encapsulates both acoustic and semantic information within a single latent space.
This entanglement prevents the model from isolating the stylistic information from the speaker's vocal characteristics when tags are simply prepended to the input prompt.

Similar solutions, such as those proposed by \citep{zhang2025minimax, du2024cosyvoice}, use a speaker embedding model to encode the speaker's voice timbre, thereby explicitly disentangling acoustic and semantic prompting.
This design simplifies the training objective, enabling the model to focus on interpreting style tags without the concurrent task of precisely cloning the reference audio but has the drawback of requiring the training of an additional speaker embedding model.

To overcome this, we constructed a dataset by pairing neutral and stylized utterances from the same speaker.
For each pair, the corresponding transcripts were concatenated, using the style tag as a delimiter, while the audio files were joined with a 0.5 to 1.5-second silence gap.
To augment the data, each neutral utterance was paired with 1 to 5 unique stylized utterances.
The dataset also included non-verbal vocalizations, which appeared in $\sim$20\% of samples, and retained $\sim$30\% of samples as unpaired neutral utterances to preserve fundamental speech synthesis capabilities.

\begin{figure}[htbp]
  \centering
  \includegraphics[width=0.8\linewidth]{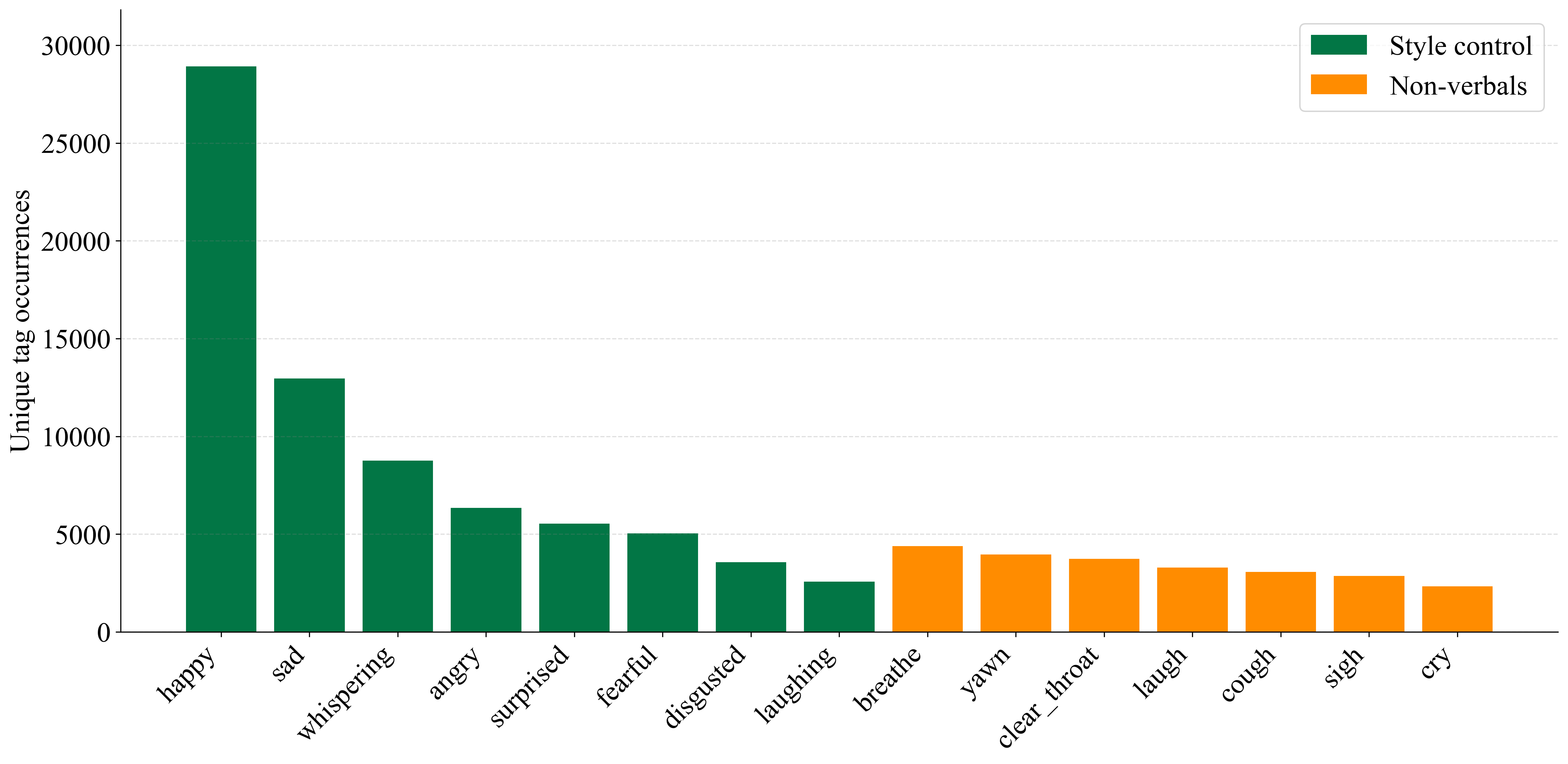}
  \caption{Histogram of occurrences of each audio markup tag.}
  \label{fig:tags_histogram}
\end{figure}

The resulting dataset, composed entirely of English audio and containing approximately 100000 examples, totals 180 hours of audio from 340 unique speakers.
A detailed distribution of audio markup tag occurrences is provided in Figure \ref{fig:tags_histogram}.

\textbf{Experiment setup.} A series of ablation studies revealed that both TTS-1 and TTS-1-Max SpeechLMs achieve better generalization, as indicated by lower evaluation loss, when fine-tuned using Low-Rank Adaptation (LoRA) \citep{hu2022lora} as opposed to full-model SFT.
The hyperparameters for LoRA fine-tuning are detailed in Table \ref{tab:audio_markup_hparams}.

\begin{table}[htbp]
  \centering
  \begin{threeparttable}
    \begin{tabular}{lcc}
      \toprule
      \textbf{Hyperparameter} & \textbf{TTS-1} & \textbf{TTS-1-Max} \\
      \midrule
      Per-GPU batch size & 4 & 2 \\
      Gradient accumulation steps & 1 & 2 \\
      Total number of GPUs & \multicolumn{2}{c}{8} \\
      Maximum sequence length & \multicolumn{2}{c}{2048} \\
      Optimizer & \multicolumn{2}{c}{AdamW, $\beta_1 = 0.9$, $\beta_2 = 0.95$} \\
      Learning rate (peak) & $4 \times 10^{-5}$ & $5 \times 10^{-6}$ \\
      Learning rate schedule & \multicolumn{2}{c}{Cosine decay (10\% warmup)} \\
      Weight decay & \multicolumn{2}{c}{0.1} \\
      LoRA layers & QKVO + MLP & QKVO \\
      LoRA rank & \multicolumn{2}{c}{16} \\
      LoRA alpha & \multicolumn{2}{c}{32} \\
      LoRA dropout & \multicolumn{2}{c}{0.2} \\
      \bottomrule
    \end{tabular}
    \vspace{0.1cm}
    \caption{Audio markup fine-tuning hyperparameters for TTS-1 and TTS-1-Max.}
    \label{tab:audio_markup_hparams}
  \end{threeparttable}
\end{table}

\textbf{Evaluation results.} The quantitative evaluation of prosodic and non-verbal synthesis fidelity presents a significant challenge.
Consequently, we conducted human-based preference testing to assess the model's controllability via audio markups.
These evaluations guided the selection of optimal model checkpoints through a simple voting mechanism.
During this process, we noted that the model demonstrated an emergent capability for style generalization to non-English languages, although with reduced fidelity. Furthermore, we observed that combining multiple style tags could produce highly nuanced and intricate vocal performances. Qualitative examples of these phenomena are provided on our project's demonstration page.

\subsection{Infrastructure}

This section covers high-level details about the hardware and software setup used for the training and evaluation of the models.

\textbf{Software.} For all the training and evaluations, we use Transformers 4.50.0 \citep{wolf2019huggingface},
PyTorch 2.6.0 \citep{paszke2019pytorch}, CUDA 12.4,
and FlashAttention 2.7.4 \citep{dao2023flashattention} for SpeechLM training acceleration.
For distributed multi-node/multi-GPU training, we use PyTorch Lightning 2.5.0 \citep{falcon2019pytorch}.

\textbf{Hardware.} For SpeechLM pre-training and supervised fine-tuning of both TTS-1 and TTS-1-Max, we used $\sim$2 months
of 4 NVIDIA H100 GPU nodes (32GPUs total). For audio codec, RL-alignment, audio markup, and all quality evaluations,
we used $\sim$3 months of 2 NVIDIA A100 GPU nodes (16GPUs total).

\textbf{Training data storage.} To efficiently search training examples, we add indexes stored alongside the audio codes.
Even with this overhead, we have achieved a compression ratio of $\sim$500:1 vs storing raw audio.
This significantly reduced storage requirements for training experiments spanning months and utilizing terabytes of distributed storage.
Moreover, this reuse of pre-tokenized audio codes for multiple training runs also accelerated the training throughput by alleviating I/O bottlenecks.

\section{Quality Evaluation}
\label{sec:quality_evaluation}

To evaluate the final performance of Inworld TTS-1 and TTS-1-Max, we generated a benchmark dataset using Gemini 2.5 Pro, comprising 100 sentences for each of the 11 supported languages, for a total of 1100 samples. We then synthesized speech for this dataset using both models with the same set of speakers to ensure a fair comparison. The quality was assessed using the same Word Error Rate (WER) and Speaker Similarity (SIM) metrics described previously.

\begin{figure}[H]
  \centering
  \includegraphics[width=0.95\linewidth]{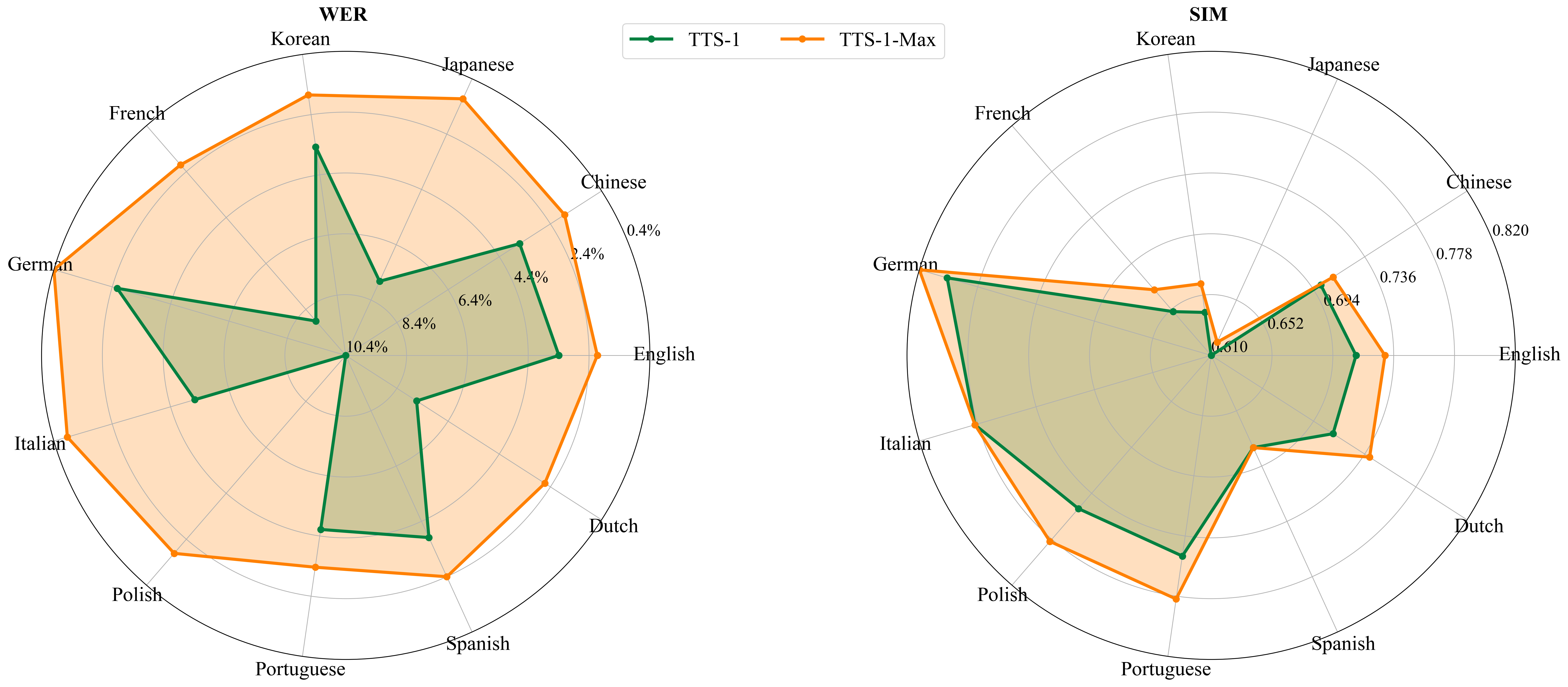}
  \caption{Multilingual evaluation results comparing WER and SIM scores by language. Left: WER (lower is better). Right: SIM (higher is better).}
  \label{fig:multilingual_eval}
\end{figure}

The results, presented in Figure \ref{fig:multilingual_eval}, show that TTS-1-Max consistently outperforms TTS-1 in terms of both WER/CER and SIM across all languages, demonstrating the effectiveness of the larger model in generating more accurate and higher-fidelity speech. Notably, both models achieve very high speaker similarity, indicating their robustness in voice cloning.

\subsection{Evaluation on Varying Input Lengths}

To further assess the models' performance on English, we conducted an evaluation across three datasets with varying input lengths: Short, Medium, and Long. The statistics for these datasets are detailed in Table \ref{tab:input_length_stats}. This evaluation aimed to measure the accuracy and robustness of the models when handling texts of different complexities and durations.

\begin{table}[htbp]
  \centering
  \begin{tabular}{lcccc}
    \toprule
    \textbf{Dataset} & \textbf{Total} & \textbf{Avg Chars} & \textbf{Min Chars} & \textbf{Max Chars} \\
    \midrule
    Short & 208 & 7.0 & 3 & 12 \\
    Medium & 150 & 87.0 & 41 & 196 \\
    Long & 517 & 153.0 & 75 & 225 \\
    \bottomrule
  \end{tabular}
  \vspace{0.1cm}
  \caption{Statistics of English evaluation datasets by input text length.}
  \label{tab:input_length_stats}
\end{table}

The evaluation results, shown in Table \ref{tab:wer_by_length}, confirm that TTS-1-Max consistently achieves lower Word Error Rates (WER) across all input lengths compared to TTS-1. This indicates that the larger model is more robust in handling a variety of text lengths, from short phrases to longer sentences. The table also includes the SFT-only results to demonstrate the improvements achieved through our complete training pipeline including RL alignment.

\begin{table}[htbp]
  \centering
  \begin{tabular}{lcccc}
    \toprule
    \textbf{Model} & \textbf{Short} & \textbf{Medium} & \textbf{Long} & \textbf{Total Avg.} \\
    \midrule
    TTS-1$_{\scriptsize\text{SFT}}$ & 11.3 & 4.0 & 8.5 & 7.9 \\
    TTS-1 & 9.6 & 2.3 & 7.0 & 6.3 \\
    TTS-1-Max$_{\scriptsize\text{SFT}}$ & 10.1 & 3.3 & 7.3 & 6.9 \\
    TTS-1-Max & 8.2 & 1.9 & 5.2 & 5.1 \\
    \bottomrule
  \end{tabular}
  \vspace{0.1cm}
  \caption{Word Error Rate (WER) in \% on English datasets of varying lengths. Models with $_{\text{SFT}}$ subscript show performance after supervised fine-tuning only, while the base models include the complete training pipeline with RL alignment.}
  \label{tab:wer_by_length}
\end{table}

\subsection{Internal TTS Arena Evaluation}

To evaluate our models against commercial TTS systems, we built an internal TTS arena for comparative evaluation. We included five TTS models in our comparison: 11LABS Multilingual V2, Cartesia Sonic 2, OpenAI TTS-1-HD, Inworld TTS-1, and Inworld TTS-1-Max.

The evaluation process involved approximately 20 human annotators conducting blind preference comparisons. The voting process was structured as follows:
\begin{enumerate}
  \item Randomly select two models from the five-model list.
  \item For each model, randomly select the corresponding built-in English speaker.
  \item Human annotators enter text to synthesize, generate audio for each pair, and vote for their preferred output.
\end{enumerate}

We collected over 400 votes, and the head-to-head win rates are presented in Table \ref{tab:arena_results}. The results show that our Inworld TTS-1-Max model achieved competitive performance, with win rates of 59.1\% against 11LABS, 60.9\% against Cartesia, 55.3\% against Inworld TTS-1, and 60.7\% against OpenAI TTS-1-HD. These results demonstrate the effectiveness of our proposed approach in producing high-quality speech that aligns with human preferences.

\begin{table}[htbp]
  \centering
  \footnotesize
  \begin{tabular}{lccccc}
    \toprule
    \textbf{Model} & \textbf{11labs} & \textbf{Cartesia} & \textbf{OpenAI} & \textbf{Inworld TTS-1} & \textbf{Inworld TTS-1-Max} \\
    \midrule
    11labs & - & 64.3\% (27/42) & 48.9\% (22/45) & 44.0\% (22/50) & 40.9\% (18/44) \\
    Cartesia & 35.7\% (15/42) & - & 40.4\% (21/52) & 36.8\% (14/38) & 39.1\% (18/46) \\
    OpenAI & 51.1\% (23/45) & 59.6\% (31/52) & - & 43.9\% (18/41) & 39.3\% (11/28) \\
    Inworld TTS-1 & 56.0\% (28/50) & 63.2\% (24/38) & 56.1\% (23/41) & - & 44.7\% (21/47) \\
    Inworld TTS-1-Max & 59.1\% (26/44) & 60.9\% (28/46) & 60.7\% (17/28) & 55.3\% (26/47) & - \\
    \bottomrule
  \end{tabular}
  \vspace{0.1cm}
  \caption{Head-to-head win rates from internal TTS arena evaluation. Each cell shows the win rate of the row model against the column model, with the number of wins and total comparisons in parentheses.}
  \label{tab:arena_results}
\end{table}

\section{Inference}

Inworld TTS-1 and TTS-1-Max alike support 2 modes of execution:
\begin{itemize}
  \item \textbf{Instant voice cloning} - the model uses only the reference audio and its transcript to produce new generations.
  \item \textbf{Professional voice cloning} - the model's SpeechLM is LoRA fine-tuned on the given user's voices recordings to further enhance similarity to the reference speaker.
\end{itemize}

Both execution modes can operate within our streaming inference pipeline, which is designed for real-time speech synthesis. The pipeline generates streaming audio segments through a careful orchestration of the SpeechLM and audio decoder. The process begins with the SpeechLM autoregressively generating audio tokens, which accumulate in a buffer. Once a buffer of a specified size is formed, the audio decoder converts this tokenized representation into 48 kHz waveform segments for delivery to the user. The following sections detail the solutions we engineered in pursuit of optimized responsiveness.

\begin{figure}[htbp]
  \centering
  \includegraphics[width=0.8\linewidth]{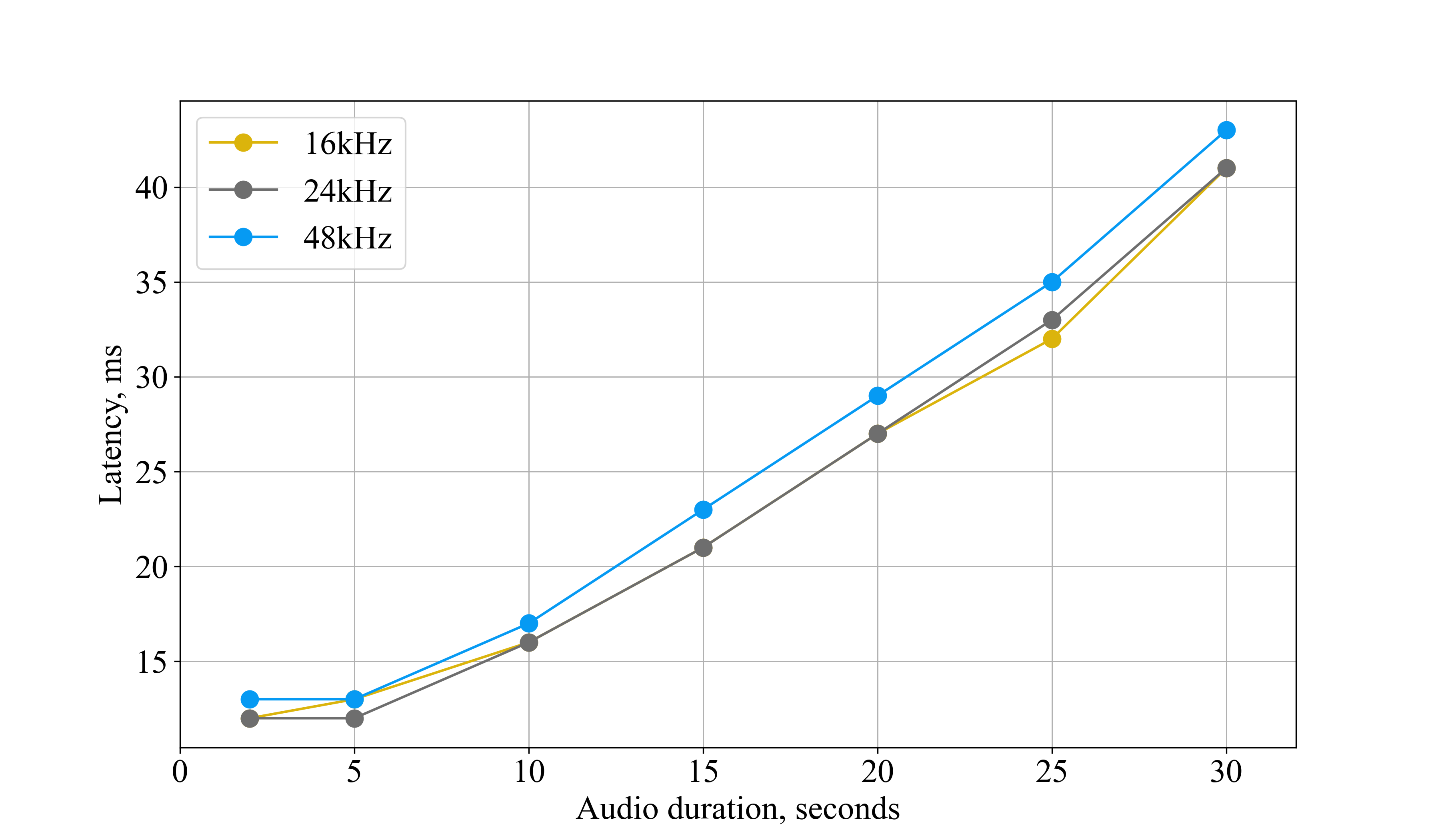}
  \caption{Audio decoder latency comparison across different sample rates and audio durations.}
  \label{fig:decoder_latency}
\end{figure}

\subsection{Decoder}

While streaming support helped to reduce the time to get the first audio chunk, it also introduced new challenges for the decoder. Specifically, the decoder must generate audio waveforms in a manner that transitions smoothly during concatenation. We observed that artifacts emerge when trying to concatenate audio segments with slight discontinuities \citep{du2024cosyvoice2}, and the inconsistency in the volume of the generated audio segments can lead to a noticeable drop in the quality of the speech. We combine the following techniques to address these issues.

\paragraph{Concatenation at the non-voicing regions.} To mitigate the artifacts caused by the audio discontinuity \citep{du2024cosyvoice2}, we propose a simple yet effective voice streaming method that only allows the audio segments to be concatenated at the non-voicing regions (e.g., a short silence between the words). When a new audio segment $\hat{X}_{t_{1}:t_{2}}$ is decoded, with $t_{1}$ and $t_{2}$ being the start and end time of the segment, we search for the last fixed-radius non-voicing region $\hat{X}_{t^{*} - \Delta t:t^{*} + \Delta t}$ that satisfies

\begin{equation}
  \label{eq:non_voicing_region}
  \max{|\hat{X}_{{t^{*} - \Delta t}:t^{*} + \Delta t}|} < \epsilon,
\end{equation}
where $\Delta t$ is the radius of the non-voicing region, and $\epsilon$ is the threshold for the signal magnitude.
If such a non-voicing region exists, the decoded audio waveform up to $t^{*}$ is emitted to the output stream, then the remaining audio waveform $\hat{X}_{t^{*}:t_{2}}$ is discarded. The tokens that correspond to the discarded audio are retained and will be processed in the next chunk. Similarly, if no region that satisfies Equation \ref{eq:non_voicing_region} is found, we discard the entire segment $\hat{X}_{t_{1}:t_{2}}$, deferring all the audio tokens for subsequent processing.

\paragraph{Volume stabilization.} Based on the observation of increased volume inconsistency in short audio segments, we propose to stabilize the volume of the generated audio by extending the context of the audio decoder. Specifically, to generate audio segment $\hat{X}_{t_{1}:t_{2}}$, we feed the audio decoder with tokens that correspond to $\hat{X}_{t_{1} -\Delta T:t_{2}}$, where $\Delta T$ is a fixed-length context. Then we remove $\hat{X}_{t_{1} -\Delta T:t_{1}}$ to obtain the desired output. Given the fact (in Table.~\ref{fig:decoder_latency}) that audio decoding is much faster than audio token generation, this method removes audible volume inconsistency with negligible latency overhead.

\paragraph{Decoding the prompt audio.} We also observed that extending the context of the audio decoder to include the prompt audio can further improve the model's ability to replicate speaker identity.
Results in Table. \ref{tab:decode_with_prompt} show that, in the non-streaming setting, the similarity scores between the prompt and the generated audio are higher if the decoder is fed with the concatenation of both the prompt audio tokens and the generated audio tokens.
In the streaming setting, this approach also benefits initial segments of the generated audio, through the extended context of the audio decoder.

\begin{table}[htbp]
  \centering
  \begin{tabular}{lc}
    \toprule
    \textbf{Setting} &  \textbf{SIM}\\
    \midrule
    Decode without prompt & 0.495 \\
    Decode with prompt & 0.535 \\
    \bottomrule
  \end{tabular}
  \vspace{0.1cm}
  \caption{Speaker similarity comparison between the audio prompt and the generated audio in the non-streaming setting. Results are computed on the English subset of the Seed-TTS evaluation \citep{anastassiou2024seed}.}
  \label{tab:decode_with_prompt}
\end{table}

\subsection{Architecture Optimizations}

To further improve latency, we collaborated with the Modular team to optimize the inference engine, resulting in several key architectural and kernel-level improvements.

\begin{itemize}
  \item The serving pipeline employs a multi-step scheduler for asynchronous speech token generation, allowing the SpeechLM to produce a batch of tokens entirely on the GPU without CPU interruption, which significantly enhances GPU utilization.
  \item The audio decoder was re-architected to support batched inputs with padding, using custom kernels to ensure correct output and process multiple sequences simultaneously.
  \item For high-performance penalty sampling, token occurrence metadata is stored in a sparse format to avoid the overhead of large tensors. The entire penalty and sampling process is executed on the GPU through highly optimized kernels written in the Mojo programming language.
  \item The inference stack leverages a graph compiler (MAX pipeline) for optimizations like kernel fusion and memory planning, complemented by custom kernels for critical operations like attention and matrix-vector multiplication, which were also developed in Mojo to outperform standard library implementations.
\end{itemize}

\begin{figure}[H]
  \centering
  \includegraphics[width=0.8\linewidth]{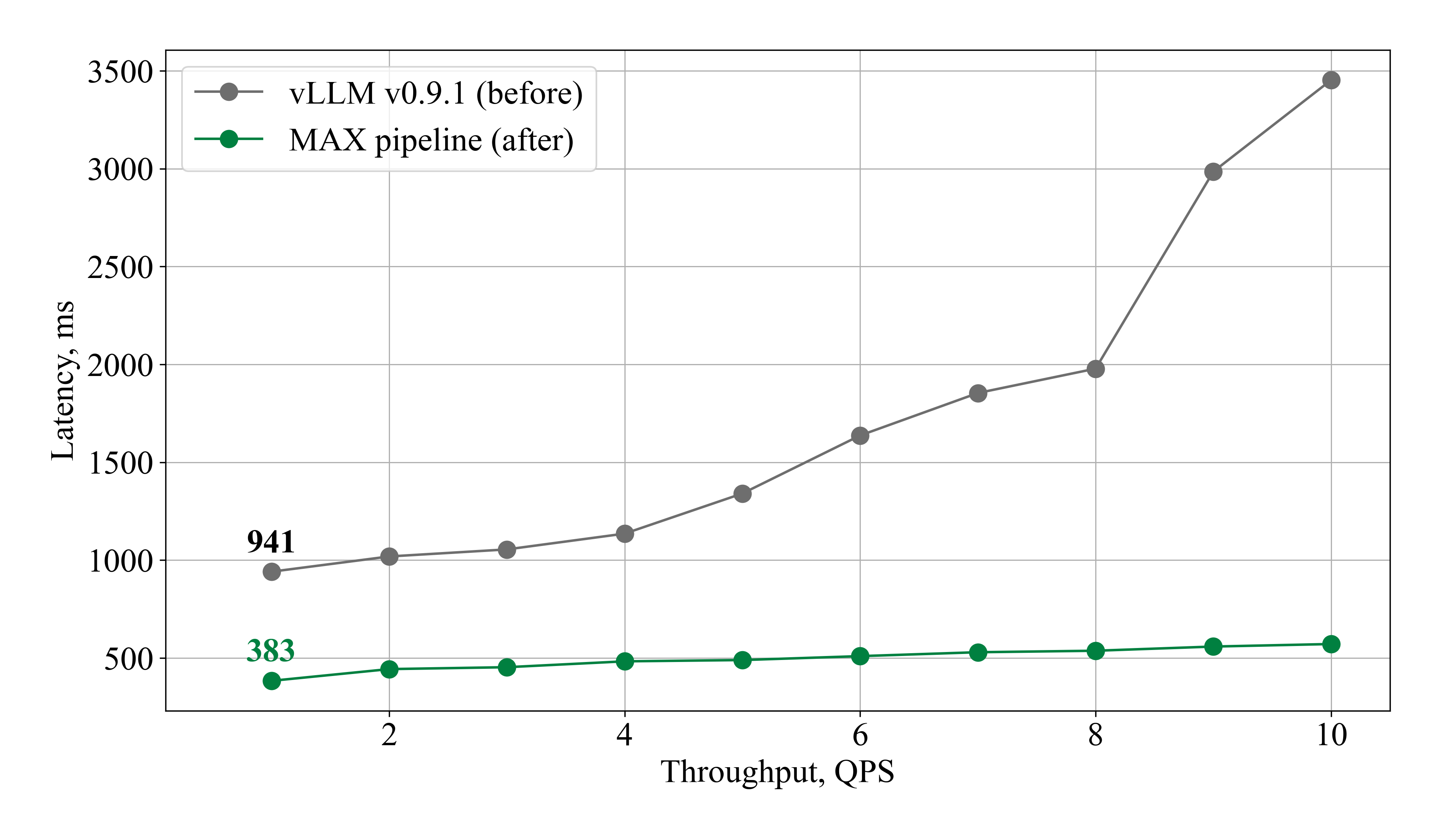}
  \caption{P90 latency comparison for the first 2-second audio chunk between the vanilla vLLM solution and the Modular-optimized implementation.}
  \label{fig:latency}
\end{figure}

As a result of these combined optimizations, the streaming API delivers the first two seconds of synthesized audio on average ~70\% faster than a vanilla vLLM-based implementation \citep{kwon2023efficient}, as shown in Figure \ref{fig:latency}.

\subsection{Model Constraints and Trade-Offs}

While the proposed inference architecture is highly efficient, it has several inherent limitations.

A key optimization is caching reference prompt audio tokens to improve latency.
However, this can cause emotional and stylistic characteristics to bleed from the reference audio into the generated speech, making it difficult to disentangle speaker identity from the reference prosody.

Furthermore, the duration of the reference audio constrains the maximum length of reliable output; longer sequences generated from short prompts may suffer from quality degradation.

The autoregressive nature of the SpeechLM necessitates a trade-off in decoding parameters.
For example, a lower sampling temperature can improve speaker similarity but reduce expressiveness.
As a result, inference parameters must be tuned for specific use cases.

Finally, imbalances in the training data affect generalization, leading to varied quality across languages and inconsistent effectiveness of audio markup tags.

\section{Ethical Considerations and Safeguards}

The development of powerful generative models for speech synthesis, particularly those with zero-shot voice cloning capabilities, raises important ethical considerations. The potential for misuse of such technology, such as the creation of synthetic media for malicious purposes, requires a proactive approach to safety. Accordingly, we have chosen not to publicly release the model weights to mitigate these risks.

In addition, all audio generated for production applications is integrated with an inaudible watermark. This technique provides a mechanism for content authentication and helps in distinguishing synthetic speech from human speech, further addressing concerns about potential misuse.

As an additional safeguard, users must explicitly confirm they have the rights to any voice they intend to clone using the instant voice cloning feature in the Inworld TTS Portal.

\section{Conclusion}

In this paper, we introduced Inworld TTS-1 and TTS-1-Max, two LLM-based text-to-speech models that demonstrate exceptional quality, controllability, and multilingual support.
Our three-stage training process combining large-scale pre-training, supervised fine-tuning, and reinforcement learning alignment has proven effective in developing high-performance TTS systems.
The development of a novel high-resolution audio codec and careful instruction data curation are key contributions that enable the generation of 48~kHz speech with fine-grained control over emotion and non-verbal vocalizations.

Despite this significant progress, our work also highlights several areas for future improvement.
We believe that more thorough and systematic data preparation and cleaning are crucial for pushing the boundaries of speech synthesis quality.
While our filtering pipeline was effective, developing more sophisticated data curation techniques could lead to even cleaner and more diverse training sets, further reducing synthesis artifacts and improving overall naturalness.

Furthermore, our evaluations, while comprehensive, reveal limitations in capturing the nuances of real-world performance.
For instance, we observed that metrics like speaker similarity can fluctuate with emotionally expressive speech delivery, suggesting that current automated evaluation protocols may not fully capture perceptual quality in dynamic, real-world scenarios.
Future work should focus on developing more robust evaluation methodologies that better account for prosodic and emotional variations, bridging the gap between quantitative metrics and human perception.

By open-sourcing our training, modeling, and benchmarking code, we hope to encourage the community to build upon our work, address these challenges, and contribute to the development of more robust, expressive, and natural-sounding speech synthesis technologies.

\clearpage

\section*{Credit Attribution and Acknowledgements}

Please cite this work as "Inworld AI (2025)".

{\scriptsize
  \begin{multicols}{2}
    \setlength{\parskip}{1ex} % Adds space between entries

    \vspace{0.4cm}
    \centerline{\normalsize\bf Training Infrastructure}

    \textbf{Compute cluster setup} \\
    Igor Poletaev, Zhifeng Deng

    \textbf{Distributed training implementation} \\
    Igor Poletaev, Yufei Feng, Feifan Fan, Phillip Dang

    \textbf{Training speed optimizations} \\
    Igor Poletaev

    \vspace{0.4cm}
    \centerline{\normalsize\bf Audio Codec}

    \textbf{Audio data collection} \\
    Yufei Feng, Igor Poletaev

    \textbf{Training pipeline implementation} \\
    Yufei Feng

    \textbf{Training run babysitting} \\
    Yufei Feng

    \vspace{0.4cm}
    \centerline{\normalsize\bf Pre-Training}

    \textbf{Audio data collection} \\
    Phillip Dang, Igor Poletaev

    \textbf{Text data collection} \\
    Igor Poletaev

    \textbf{Training pipeline implementation} \\
    Igor Poletaev

    \textbf{Training run babysitting} \\
    Igor Poletaev

    \vspace{0.4cm}
    \centerline{\normalsize\bf Fine-Tuning}

    \textbf{Audio data collection} \\
    Igor Poletaev, Pavel Filimonov, Phillip Dang, Yufei Feng, Feifan Fan, Michael Ermolenko

    \textbf{Data annotation and filtering} \\
    Igor Poletaev, Pavel Filimonov, Phillip Dang, Yufei Feng, Feifan Fan, Michael Ermolenko, Pavel Karpik

    \textbf{Optimizations and architecture} \\
    Igor Poletaev, Yufei Feng, Michael Ermolenko

    \textbf{Training runs babysitting} \\
    Igor Poletaev

    \vspace{0.4cm}
    \centerline{\normalsize\bf RL Alignment}

    \textbf{Audio data collection} \\
    Feifan Fan, Igor Poletaev

    \textbf{Data annotation and filtering} \\
    Feifan Fan, Igor Poletaev

    \textbf{Training pipeline implementation} \\
    Feifan Fan

    \textbf{Training runs babysitting} \\
    Feifan Fan

    \vspace{0.4cm}
    \centerline{\normalsize\bf Quality Evaluations}

    \textbf{Benchmarking data collection} \\
    Feifan Fan, Igor Poletaev

    \textbf{Benchmarking infrastructure} \\
    Feifan Fan, Igor Poletaev, Phillip Dang, Yufei Feng

    \vspace{0.4cm}
    \centerline{\normalsize\bf Production Serving}

    \textbf{Compute cluster setup} \\
    Zhifeng Deng, Oleg Atamanenko, Nurullah Morshed

    \textbf{Deployment} \\
    Zhifeng Deng, Pavel Karpik, Feifan Fan, Oleg Atamanenko, Nurullah Morshed

    \textbf{Architecture optimizations} \\
    Pavel Karpik, (Tyler Kenney, Brian Zhang, Austin Doolittle, Shouzheng Liu, Hengjie Wang, Chris Elrod)\footnotemark

    \textbf{Text pre-processing and normalization} \\
    Pavel Filimonov, Pavel Karpik

    \textbf{Audio post-processing} \\
    Yufei Feng, Pavel Karpik

    \textbf{Prebuilt voice selection} \\
    Robert Villahermosa, Jean Wang, Phillip Dang, Igor Poletaev

    \textbf{Quality assurance} \\
    Anna Chalova, Robert Villahermosa, Jean Wang, Phillip Dang

    \textbf{Audio watermarking} \\
    Pavel Karpik

    \textbf{Voice cloning infrastructure} \\
    Jimmy Du, Vikram Sivaraja, Rinat Takhautdinov, Peter Skirko, Oleg Atamanenko, Mikhail Mamontov, Feifan Fan

    \textbf{Inworld TTS Portal UI \& UX} \\
    Jasmine Mai, Dmytro Semernia, Joseph Coombes, Jean Wang, Suri Mao, Valeria Gusarova

    \textbf{Public API} \\
    Ian Lee, Jimmy Du, Pavel Karpik, Igor Poletaev

    \vspace{0.4cm}
    \centerline{\normalsize\bf Additional Contributions}

    \textbf{Blog post \& paper content} \\
    Igor Poletaev, Feifan Fan, Yufei Feng, Evgenii Shingarev, Joseph Coombes

    \textbf{Launch partners} \\
    Cheryl Fichter, Nikki Cope, Louis Fischer

    \textbf{Legal} \\
    Oliver Louie

    \textbf{System administration \& on-call support} \\
    Oleg Atamanenko, Nurullah Morshed, Evgenii Shingarev

    \textbf{Communications} \\
    Kylan Gibbs, Florin Radu, Cheryl Fichter, Andreas Assad Kottner

    \textbf{Technical report} \\
    Igor Poletaev, Feifan Fan, Yufei Feng, Phillip Dang, Pavel Karpik

    \textbf{Authorship \& credit attribution} \\
    Igor Poletaev

  \end{multicols}
}

\footnotetext{\href{https://modular.com/}{Modular Team}}

We thank all Inworld AI team members for their contributions, including those not explicitly named above.
These results would not have been possible without our collective effort.

We also thank our partners: Oracle and the OCI HPC team for infrastructure support, Voltage Park for providing H100 GPUs, and the amazing Modular team for optimizing model serving and making its real-time inference ready.

\clearpage

\bibliography{references}
\bibliographystyle{unsrtnat}

\end{document}